\def\year{2021}\relax
\documentclass[letterpaper]{article} 
\usepackage{aaai21}  
\usepackage{times}  
\usepackage{helvet} 
\usepackage{courier}  
\usepackage[hyphens]{url}  
\usepackage{graphicx} 
\usepackage{graphicx}  
\usepackage{natbib}  
\usepackage{caption} 
\nocopyright
\usepackage[misc]{ifsym}

\title{
    BGM: Building a Dynamic Guidance Map \\without Visual Images for Trajectory Prediction
}
\author{
    Beihao Xia\textsuperscript{*},
    Conghao Wong\footnote{indicates equal contribution.},
    Heng Li,\\
    Shiming Chen,
    Qinmu Peng,
    Xinge You \textsuperscript{\Letter} \\
}
\affiliations {
    School of Electronic Information and Communications, \\Huazhong University of Science and Technology, Hubei, China\\
    \{xbh\_hust, conghao\_wong, liheng, shimingchen, pengqinmu, youxg\}@hust.edu.cn
}

\usepackage{multirow}
\usepackage{amsmath, amsfonts}
\usepackage[ruled,vlined]{algorithm2e}

\begin{document}
\maketitle

\begin{abstract}
Visual images usually contain the informative context of the environment, thereby helping to predict agents' behaviors.
However, they hardly impose the dynamic effects on agents' actual behaviors due to the respectively fixed semantics.
To solve this problem, we propose a deterministic model named BGM to construct a guidance map to represent the dynamic semantics, which circumvents to use visual images for each agent to reflect the difference of activities in different periods.
We first record all agents' activities in the scene within a period close to the current to construct a guidance map and then feed it to a Context CNN to obtain their context features.
We adopt a Historical Trajectory Encoder to extract the trajectory features and then combine them with the context feature as the input of the social energy based trajectory decoder, thus obtaining the prediction that meets the social rules.
Experiments demonstrate that BGM achieves state-of-the-art prediction accuracy on the two widely used ETH and UCY datasets and handles more complex scenarios.
\end{abstract}

\section{Introduction}
Trajectory prediction is to forecast agents' locations in the future based on their past positions.
It is popular and widely applied in self-driving \cite{desire,deo2018would,rhinehart2019precog,trafficPredict}, robotic navigation \cite{unfreezing}, tracking \cite{learningSocialEtiquette} and more other areas.
Many researchers are committed to this topic and made their contribution.
Early works like \cite{socialLSTM,cidnn,socialAttention} focus on predicting agents' future trajectories by just modeling social interaction.
In addition to being affected by others, agents always consider planning their future routes under environmental constraints.
Works like \cite{ssLSTM,sophie,bigat} extract visual features of the scenes by CNNs to help predict agents' future trajectories, which shows their effectiveness.
The CNN features represent the scene environment around agents so that models could learn agents' activities in different scene characteristics, to give better predictions in other scenes with similar looks.
More additional inputs like scene segmented images and person behaviors are also adopted in \cite{peekingIntoTheFuture} to describe the agents' past actions and their corresponding context comprehensively to achieve better results.

\begin{figure}[t]
\centering
\includegraphics[width=0.96\columnwidth]{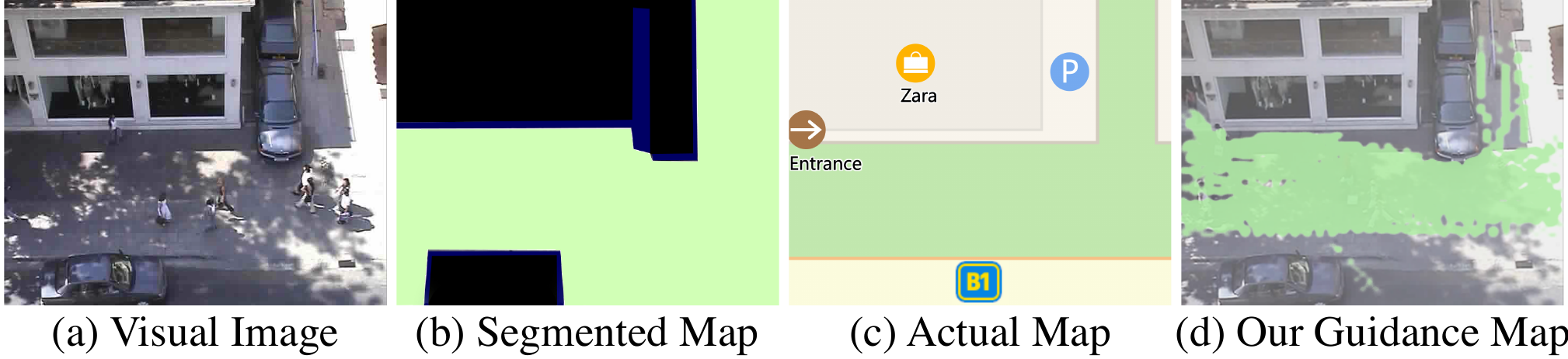}
\caption{
Different maps to describe the same scene.
(a) an original visual scene image from the UCY-Zara1, (b) the corresponding segmented map, (c) the actual map, and (d) the dynamic guidance map constructed by BGM.
Green parts in the above maps indicate the areas where agents could walk.
}
\label{fig_introduction}
\end{figure}

Most existing methods like \cite{manh2018scene,sophie,peekingIntoTheFuture,bigat} tackle agents' movement and scene context separately.
They do not consider the impact of real agents' activities and the dynamic changes of their preferences in the scene when modeling the agent-scene interaction.
An easily overlooked fact is that the scene interaction is not static as time goes by.
Several researchers treat the scene interaction as the influence of static background objects (such as trees, roads, and buildings) on agents.
How to describe the interaction between a standing pedestrian and another moving one?
Previous methods usually consider it as the agent-agent interaction.
However, it is incredibly similar to the impact of trees on other agents since they both hinder others' movements.
It means that the interaction among different agents and the scene are connected rather than separate.
Thus, it is not comprehensive that only use the visual images or static segmented images to model the generally talked scene interaction.
We regard the factors that may affect the scene's interactive behaviors as a whole and summarize it as the scene context.
The scene context contains both dynamic factors and interaction producers, which have no categories limitations.
For instance, standing pedestrians and parking cars are part of the scene context, similar to trees and stores' doors.

As shown in Fig.\ref{fig_introduction}, we offer different maps that gather the scene context.
However, (i) Static visual images cannot describe the dynamics of the scene, since the flow of agents and their interaction behaviors could change over time;
(ii) The segmented map only shows the rough scene annotations.
However, it does not point out the interactive components like Zara's entrance as the actual map does.
The agent's historical trajectory is the reflection of his past activities \cite{hong2020heteta}.
Considering the disadvantages of these maps, we propose a novel dynamic guidance map to model the changeable scene context.
We select a proper time window to record available trajectories in the scene and then predict agents' future trajectories with the help of the distribution of observed trajectories near the agent, thus reflecting the scene context's dynamics.
The guidance map contains the dynamic change of walking areas and interaction preferences, as well as interactive components like the entrance of shops, which static visual images and segmented images do not contain.

To better model social interaction, most current methods judge each agent's contribution based on their relative distance among the participators.
However, they cannot well model the interaction situation when the distances are the same.
For example, when a pedestrian is running, and another is walking towards another same pedestrian at the same distance.
These models cannot better describe this case since the two people have different contributions.
Besides, previous methods also limit the number of agents participating in the interaction due to their network structures.
Inspired by the work\cite{socialForce,whoAreYou}, we build a social energy field and regard that the agents will always pass through lower social energy areas.
We design a social module to make the model's outputs socially acceptable based on the social energy field, which needs not limit the number of agents participating in the interaction.
Besides, the universal form of energy function makes the model more robust to various social situations.

The main contributions of this work are listed as follows:
\begin{itemize}
    \item A guidance map with more semantics is constructed better to predict future trajectories in response to the dynamic context.
    \item A social module is designed to make the predictions in line with social etiquette.
    \item Experiments show that our BGM has state-of-the-art prediction performance on ETH and UCY datasets compared with other models.
\end{itemize}

The rest of the paper is organized as follows:
We give a brief overview of related work in Section 2.
In Section 3, we will give a detailed description of our model.
We show the performance of BGM in Section 4.

\section{Related Work}
\subsection{Trajectory Clustering Analysis}
Finding a suitable representation of historical trajectories is a significant step for trajectory prediction.
It is essential to study the task of trajectory data mining.
Agglomerative clustering model \cite{zheng2009mining} is explored to mine the interesting locations to catch trajectories.
Considering different lengths of trajectories, the work \cite{2010Object} construct the affinity matrix.
Some spectral clustering models like \cite{2016ReD} are proposed to capture causal relationships between time series.

\subsection{Description of Social Interaction}
The earliest work Social Force Model \cite{socialForce} describes the social interaction with speed, acceleration, direction, and other social force items.
In addition to these factors, later works like \cite{activityForecasting,youWillNeverWalkAlone} add different aspects to model interaction among agents.
With the development of data-driven methods, Social LSTM \cite{socialLSTM}, which utilizes LSTM \cite{lstm} and an additional social pooling layer to extract social features, is proposed to learn to predict human behaviors in crowded scenes.
Besides, works like \cite{fernando2018soft+,socialAttention} take advantage of the attention mechanism to judge each agent's contribution to the interaction considering the different influences between varied agents and the target one.
Due to graph neural networks' success, some methods based on graph attention networks like \cite{stgat,bigat,Sun_2020_CVPR} also try to model social interaction with the graph network dynamically.
However, these methods have limitations on the number of nodes in the graph, limiting the number of agents participating in the interaction.
Works like \cite{cidnn,sophie} define a maximum number of agents participating in the interaction.
Others like \cite{socialGAN} take a spatial window to obtain the neighbors of the current agent and model interactions among them, which may ignore the further agent with a more outstanding contribution.
To deal with all possible social interaction, the approach \cite{multiAgentTensorFusion} construct a new tensor, which encodes trajectories of all agents in the crowed scene.
Unlike other works, \cite{traphic} and \cite{trafficPredict} also considered the heterogeneity of agents caused by the different categories of them.

\subsection{Description of Scene Interaction}
The scene context would also change agents' future movements.
Work \cite{learningSocialEtiquette} annotates scenes with static semantic segmentation results since they regard that people always walk under the rules of common sense.
Some works \cite{sophie,bigat} use Convolutional Neural Networks to extract the visual features.
With the development of semantic segmentation, more detailed extra information are available in recent works like \cite{peekingIntoTheFuture,manh2018scene,ssLSTM}.
What is more, agents' physical activities are used in \cite{peekingIntoTheFuture} to make a better prediction.
However, it should be noted that these methods that obtain scene information with the help of additional tools make it lose the connection between the scene and agents, resulting in the inability to make full use of the auxiliary information when making predictions.

\section{Model}
\begin{figure*}[t]
\centering
\includegraphics[width=2.06\columnwidth]{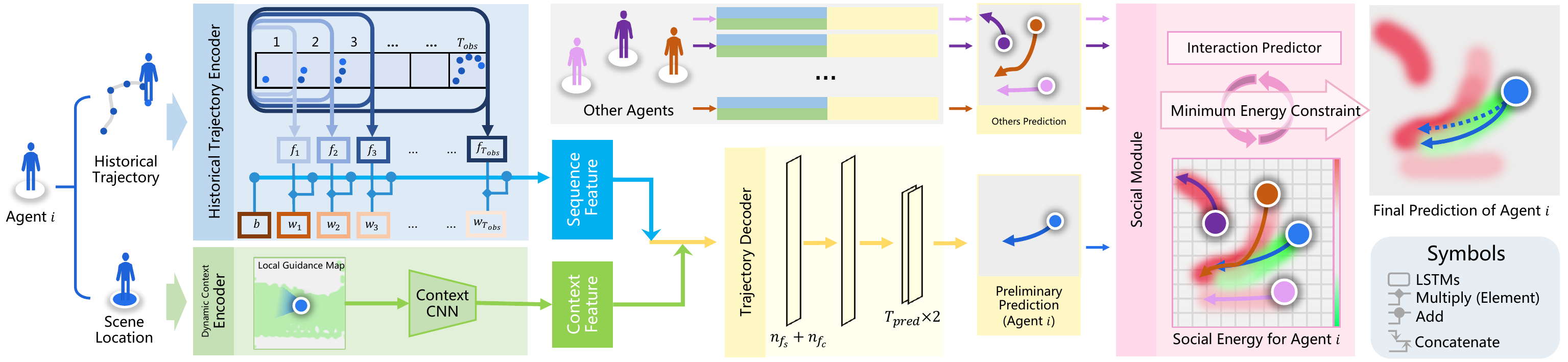}
\caption{Structure of BGM.
The BGM contains three main components: (i) a Historical Trajectory Encoder, (ii) a Dynamic Context Encoder and (iii) a Social Energy based Trajectory Decoder, which includes a Trajectory Decoder and a Social Module.}
\label{fig_model}
\end{figure*}

\subsection{Problem Definition}
Given a sequence of observed historical trajectory of $i$-th agent from time step $T_0$ to $T_0+T_{obs}-1$, the prediction task is to find a possible future trajectory from time step $T_0+T_{obs}$ to $T_0+T_{obs}+T_{pred}-1$.
$T_{obs}$ and $T_{pred}$ represent the length of observations and predictions respectively.
To describe more clearly, we define $T_1 = T_0 + T_{obs}$ and $T_2 = T_1 + T_{pred}$.
Use $p^i_t = (x^i_t, y^i_t)$ to represent the coordinate of $i$-th agent at time step $t$.
Its observed trajectory is denoted by $X^i_{T_0 : T_1} = \{p^i_{T_0+t}\in \mathbb{R}^2 |\ t = 0, ..., T_{obs}-1 \}$ and ground truth trajectory by $Y^i_{T_1 : T_2} = \{p^i_{T_1+t} \in \mathbb{R}^2 |\ t = 0, ..., T_{pred}-1\}$.
The prediction problem is to find a model $F(\cdot)$ to predict the agent's reasonable future trajectory $\hat{Y}^i_{T_1 : T_2} = F(X^i_{T_0 : T_1})$.

\subsection{Overview}
As shown in Fig.\ref{fig_model}, the BGM contains three main parts: a Historical Trajectory Encoder (HTE), a Dynamic Context Encoder (DCE), and a Social Energy based trajectory Decoder (SED).
The HTE extracts multi-level trajectory features on different time scales and then employ them as a set of bases to represent the whole observed trajectory.
Simultaneously, the DSE will create a dynamic scene activity map, Guidance Map, using all observed trajectories during a specific period.
Finally, both the trajectory representation and the guidance map around one agent will be fed to the SED to generate future predictions under the social rules.

\subsection{Historical Trajectory Encoder}
We build a historical trajectory encoder to focus on the essential behaviors that may affect the future, which occurred in the distant past during the observation period.
Previous methods like \cite{socialLSTM,socialGAN} pay more attention to the state of trajectory at the last moments closer to the current due to the structure of basic LSTM.
We select several time windows with different lengths to obtain a subset of historical trajectories and then feed them to LSTMs (weights are shared) to get the trajectory status at all other observation moments.
For agent $i$, we divide his historical trajectory $X^i_{T_0 : T_1}$ in chronological order into $T_{obs}$ different scales as $\{X^i_{T_0 : T_0 + t}|\ t = 1, ..., T_{obs}\}$.
Then we input all of them to the LSTMs to obtain a set of multi-level historical trajectory features $F^i_{T_0:T_1} = \{\mbox{LSTM}[\phi(X^i_{T_0 : T_0 + t})]|\ t = 1, ..., T_{obs}\}$, where $\phi$ denotes the position embedding function.
To make better use of this set of features, we introduce a set of trainable weight matrices $W_{T_0 : T_1} = \{w_1, w_2, ..., w_{T_{obs}}\}$ and a bias vector $b$ to weight each of $F^i_{T_0:T_1}$, to obtain the final sequence feature $f_{S, T_0:T1}^i$ of agent $i$.
Formally,
\begin{equation}
    \label{eq_FC}
    f_{S, T_0:T1}^i = \sum_{t=1}^{T_{obs}} w_t \mbox{LSTM}[\phi(X^i_{T_0 : T_0 + t})] + b.
\end{equation}

\subsection{Dynamic Context Encoder}
\begin{figure}[t]
\centering
\includegraphics[width=0.8\columnwidth]{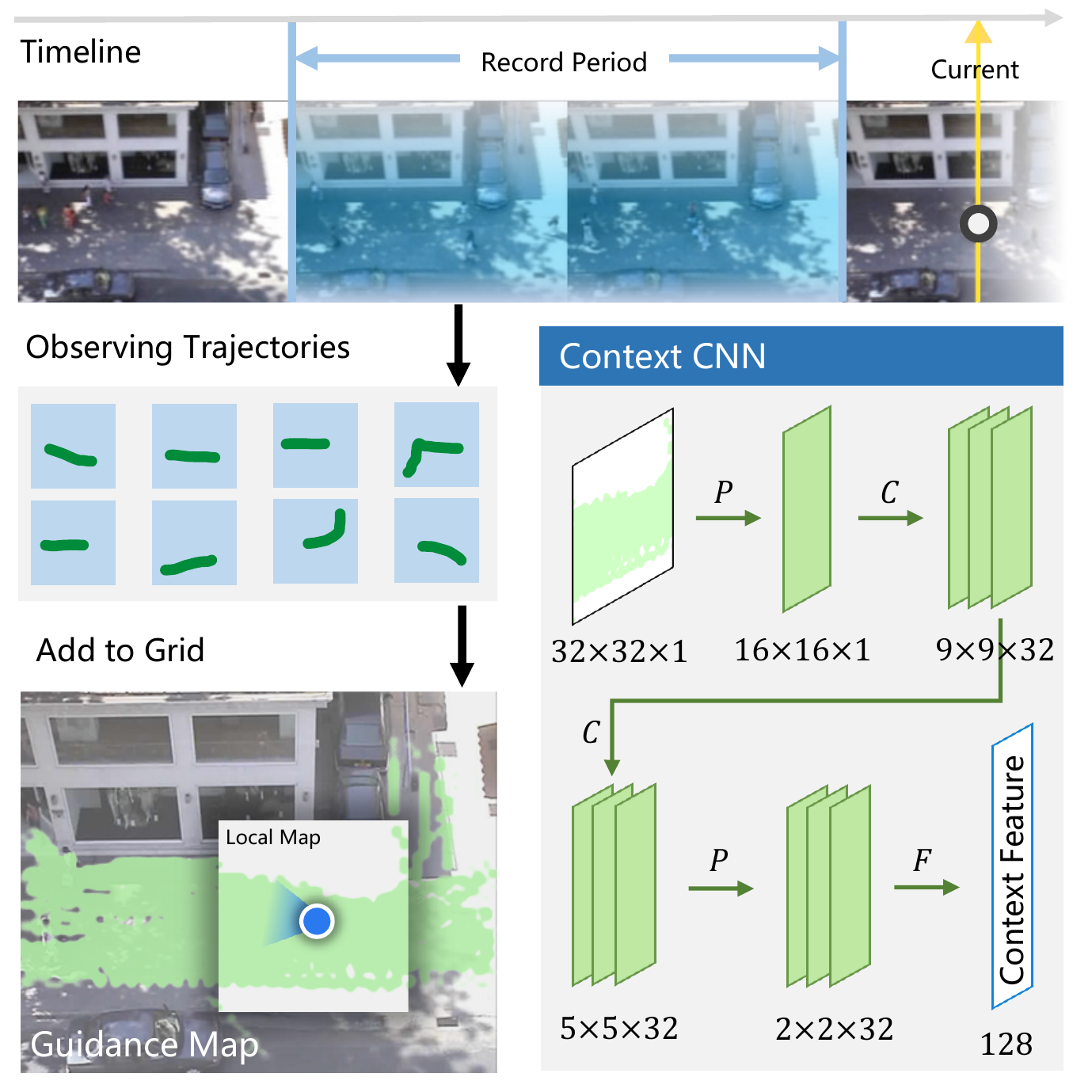}
\caption{
Structure of the Dynamic Context Encoder.
$ P $, $ C $, and $ F $ represent average pooling layers, convolution layers, and flatten operation.
The video stream shows the time course of the prediction.
We find a time window as the recording period and use all the trajectories saved to construct the complete guidance map.
We cut partial from the whole guidance map as one's local map and then input it to the context CNN to obtain his context feature.
}
\label{fig_cnn}
\end{figure}

This encoder computes a dynamic context map called Guidance Map, which displays other agents' trajectory distribution near the target agent during observation.
We use a new update strategy to maintain the time efficiency of the guidance map.
We first define the time period this encoder used to create the guidance map as the Record Period $T_R = \{t_{r1}, t_{r2}, ..., t_{rn}\}$.
We shift this time window $ T_R $ forward through training or testing to capture specific interactive cases to reflect the scene context's dynamic changes.
For agents appeared in $T_R$, we collect all of their observed positions as the Record Positions $X(T_R) = \{p_i = (x_i, y_i)|\ i\}$.
We adopt an adaptive selection strategy to make the guidance map more robust by limiting the number of records and this window's length.
We show the details in Algorithm.\ref{alg_timewindow}.

\begin{algorithm}[t]
\caption{A selection strategy of $T_R$ and $X(T_R)$.}
\label{alg_timewindow}
\KwIn{
    Video clip $V=\{V_i\}_{i=1}^{T_V}$,
    an agent detector or manual marker $Det(\cdot)$,
    max waiting frames $T_{max}$,
    minimum record limitation $N_{min}$ and maximum record limitation $N_{max}$ in the window
    and the beginning time $T_p$ for prediction ($T_p\leq T_V$).
}
\KwOut{
    Time window $T_R$ and a set of recorded positions $X(T_R)$.
}
\BlankLine
Initialize $f = 1$ and sets $T_R = X(T_R) = \emptyset$\\
Initialize temporary sets $T_R^s = X^s(T_R) = \emptyset$

\While{$f\leq T_{p}$}{
    Detecting or marking agents by: $P = \{(x_i, y_i)\}_{i=1}^{N_f} = Det(V_f)$  \\
    Update $T_R = T_R \cup f $  \\
    Update $X(T_R) = X(T_R) \cup P$    \\
    $f = f + 1$ \\

    \uIf{$(|X(T_R)|\geq N_{max})$ or $(|\ t_R|\geq T_{max}$ and $|X(T_R)|\geq N_{min})$}{
        Save current window and positions: $T_R^s = T_R,\ X^s(T_R) = X(T_R)$    \\
        Clean up sets $T_R = X(T_R)=\emptyset$
    } \ElseIf{$|\ t_R|\geq T_{max}$}{
        Clean up sets $T_R = X(T_R)=\emptyset$
    }

}

\Return{$T_R = T_R^s$, $X(T_R) = X^s(T_R)$}
\end{algorithm}

We use a 2-dimension grid $M(T_R) \in \mathbb{R}^{H \times W}$ to represent the distribution of these recorded trajectories in the scene.
Denote the mapping function from actual position $p$ to grid position $p_g$ by $p_g = m(p)$.
The complete guidance map in a specific $T_R$ is computed by:
\begin{equation}
\begin{aligned}
    M(T_R) &= \sum_{p_0 \in X(T_R)}\delta(p; m(p_0)),\\
    \mbox{where}\ \delta(p) \in &\mathbb{R}^{H \times W}\ \mbox{and}\ \delta(p; x_0) = \left\{
        \begin{aligned}
        &1, \mbox{if}\ p = x_0,\\
        &0, \mbox{otherwise}.
        \end{aligned}
        \right.
\end{aligned}
\end{equation}
For target agent $i$, we take a part of the complete map above surrounding him as his Local Guidance Map $M_L^i(T_R) = M(T_R)[m(p)-(l,l):m(p)+(l,l)]$,
where $p$ represents his last observed position, and the side length of the local guidance map is $2l$.
Then we use a Context CNN (C-CNN) (detailed structure can be seen in Fig.\ref{fig_cnn}) to obtain his final context feature.
Formally,
\begin{equation}
\label{eq_map}
    f_C^i(T_R)=\mbox{C-CNN}[M_L^i(T_R)].
\end{equation}
For the context feature $f_C^i(T_R)$, we can see that it will become different when $i$ and $T_R$ changes.
This well-designed context feature could adapt to different periods and different agents, reflecting the dynamic adaptability of our BGM.

\subsection{Social Energy based Trajectory Decoder}
A joint feature $f^i$ is concatenated by the sequence feature $f_S^i$ and context feature $f_C^i(T_R)$ above to show both historical activities and future intention of the target agent $i$ according to his local context.
(Subscript $ _{T_1: T_2}$ is omitted for a clear expression when there are no conflicts.)

We use an MLP (denoted by $\mbox{MLP}_D$) to process this joint feature to obtain the preliminary prediction, formally ${\hat{Y}_0}^i = \mbox{MLP}_D(f^i)$.
Note that the joint feature $f^i$ contains the record of all historical behaviors of agent $i$, which are reflected by his observed trajectory.
Therefore, the output dimension of $\mbox{MLP}_D$ is set to $2T_{pred}$ to obtain the entire prediction just by one step of the calculation, which reduces the cumulative effect of errors due to the circular calculations and the information omissions.
Besides, we employ an Interaction Predictor (marked with $P$) and a Social Energy based Discriminant Function (marked with $D$) to make sure that the prediction is with social etiquette.
We call the $P$ and the $D$ together as the {\bf Social Module}.

\subsubsection{Social Energy based Discriminant Function}
We use a widely applicable and reasonable energy function method to describe agents' social-behavioral preferences.
Note that we build this function for each agent, which means that each agent in the scene has his discriminant function different from others.
Define the static social energy exhibited by some other agents at position $p_o$ to the target agent as:
\begin{equation}
    f(p; p_o, r, a) = \left\{
        \begin{aligned}
            a - \frac{a}{r}\Vert p - p_o \Vert  & , & \Vert p - p_o \Vert \leq r, \\
            0 & , & \Vert p - p_o \Vert > r.
        \end{aligned}
        \right.
\end{equation}
Where $p \in \mathbb{R}^2$ indicates the location variable for current agent's energy function, $r$ and $a$ represent the range and amplitude of energy respectively.
We classify items that could affect agents' future trajectories and activities into three parts: destination, interplay, and social etiquette.
Use $\Phi = \{\lambda_d, \lambda_i, \lambda_s, r_d, r_i, r_s\}$ to denote a set of social parameters.
The social energy function for agent $i$ is defined as:
\begin{equation}
\begin{aligned}
    E^i(p; \Phi) =& \lambda_d E^i_D(p; r_d) + \lambda_i E_I^i(p; r_i) + \lambda_s E^i_S(p; r_s) \\
    =& \lambda_d \sum_{p_0 \in \hat{Y}^i_0} f(p; p_0, r_d, -1)\ + \\
    &\lambda_i \sum_{j \in J^i} \sum_{p_j \in \hat{Y}^j_0} d(i, j)v(i,j)f(p; p_j, r_i, -1)\ + \\
    &\lambda_s \sum_{j \in J^i} \sum_{p_j \in \hat{Y}^j_0} f(p; p_j, r_s, 1).
\end{aligned}
\end{equation}

(i) Destination $E^i_D(p; r_d)$.
The destination refers to the desired future trend under one's own will.
We use the preliminary prediction, $\hat{Y}^i_0$, to represent the original future trends of agent $i$.

(ii) Interplay $E_I^i(p; r_i)$.
Pedestrians always follow their companions and avoid others coming oncoming to prevent probable collisions.
Agents' avoidance behaviors have a relationship with other directions and velocities.
$J^i = \{j_1, ..., j_n\}$ represents all other agents appeared during the prediction period except agent $i$.
We use $d(i, j)$ to describe the overall direction difference for the observed trajectories of between agent $i$ and $j$, and $v(i, j)$ to describe the relative velocity.
Formally:
\begin{equation}
\begin{aligned}
    d(i, j) &= \frac{(p^i_{T_0} - p^i_{T_1 - 1})^T (p^j_{T_0} - p^j_{T_1 - 1})}{\Vert p^i_{T_0} - p^i_{T_1 - 1} \Vert \Vert p^j_{T_0} - p^j_{T_1 - 1}\Vert}, \\
    v(i, j) &= \Vert p^j_{T_0} - p^j_{T_1 - 1} \Vert / \Vert p^i_{T_0} - p^i_{T_1 - 1}\Vert.
\end{aligned}
\end{equation}

(iii) Social Etiquette.
When planning future journeys, pedestrians always avoid getting too close to others to arouse their disgust.
It is also called a safe social distance.

A position with higher social energy means that it could make others feel uncomfortable if that agent insists on passing nearby.
The Social Energy based Discriminant Function is defined as:
\begin{equation}
    D(\hat{Y}^i_0) = \sum_{p \in \hat{Y}^i_0} E^i(p).
\end{equation}

\subsubsection{Interaction Predictor}
We use an Interaction Predictor to generate a little local variation to the preliminary prediction according to the social energy function to align with the social rules.
For preliminary prediction of agent $i$ at time $t$, define the $k$ order ($k\geq1$) interaction predictor as:
\begin{equation}
    P^k(p^i_t) = \left\{
        \begin{aligned}
        p^i_t-\theta \left.\frac{\mathrm{d}E^i(p)}{\mathrm{d}p}\right|_{p=p^i_t}, \ \ k=1,\\
        P^1(P^{k-1}(p^i_t)), \ \ k\geq 2.
        \end{aligned}
        \right.
\end{equation}
$\theta$ represents the update coefficient.
To make sure that the prediction is with social etiquette, we gradually increase the order $k$ of the interaction predictor until $D$ gives a steady value.
Formally, the final prediction is obtained by:
\begin{equation}
\begin{aligned}
    \hat{Y}^i_{T_1:T_2} &= P^k(\hat{Y}^i_0) = \{P^k(\hat{p^i_t}) |\ t = T_1, ..., T_2-1\},\\
    &\mbox{where}\ |D[P^k(\hat{Y}^i_0)] - D[P^{k-1}(\hat{Y}^i_0)]| \leq \epsilon.
\end{aligned}
\end{equation}

\subsection{Loss Functions and Implementation Details}
For the historical trajectory encoder, context CNN and $\mbox{MLP}_D$,
we treat all these components as a whole end-to-end network and use the L2 loss function to train it.
For each point in preliminary prediction of $i$-th agent $p^i_t \in {\hat{Y}_0}^i$ and their ground truth ${p^i_{real}}_t \in Y^i$, the loss function can be written as:
\begin{equation}
    L = \sum_i \sum_t \Vert p^i_t - {p_{real}^i}_t \Vert.
\end{equation}
We train the network with the Adam optimizer with a learning rate of 0.01 in 500 epochs.
For the original $T_{obs}\times 2$ historical trajectories, we embed them into $T_{obs}\times 64$ vectors.
The dimensions of the hidden state of LSTM is 64.
We also use one layer of MLP with a 256-dimension's output and ReLU activation to process the sequence feature.
We set the resolution of the grid guidance map to 0.25 meters per grid.
For the current task of predicting future 4.8 seconds, we use a square window with side length $2l=8$ meters to prepare local guidance maps for each agent.
We also guide two MLPs to increase the dimension of both the sequence and the context feature into 256.
We concatenate the sequence feature and the context feature into a 512-dimension feature vector, and then feed it to the $MLP_D$ ($512\to T_{pred}\times 64 \to T_{pred}\times 2$) to get the preliminary prediction.
ReLU activations are applied in $MLP_D$ except for its last output layer.
We build each agent's energy function with a grid format like the social module's guidance map.
Its resolution is set to 0.1 meters per grid.
Due to a safe social distance is about 0.1 meters \cite{sophie}, we set social parameters as $\Phi=\{1.0, 1.0, 0.2, 2\mbox{meters}, 1.5\mbox{meters}, 0.1\mbox{meters}\}$.
The update coefficient $\theta$ is set to 0.001.
In experiments, we use a 10-order interaction predictor to make the final prediction socially acceptable.

\section{Experiments}
\subsubsection{Datasets}
We have evaluated BGM on two public human trajectory datasets: ETH \cite{whoAreYou} and UCY \cite{ucy}.
These two datasets contain five sceneries: eth, hotel, zara1, zara2, and univ.
They are annotated trajectories of pedestrians in real-world scenes.

\subsubsection{Evaluation Metrics}
Similar to previous works like \cite{socialLSTM,socialGAN,bigat}, we use the Average Displacement Error (ADE) and the Final Displacement Error (FDE) to describe the performance.
Their specific calculation method can refer to Eq.7 in \cite{cidnn}.
In particular, for models that generate multiple predictions, we use the minADE \cite{stgcnn} among $K$ trajectories to evaluate their performance.
Our experiments show the prediction performance of agents in the future 4.8 seconds (12 frames) using their trajectories during the past 3.2 seconds (8 frames).

\subsubsection{Baselines}
We choose deterministic models, like CIDNN \cite{cidnn}, SR-LSTM \cite{srLSTM}, and generative models, like Social GAN \cite{socialGAN}, STGAT \cite{stgat}, SoPhie \cite{sophie}, Social-BiGAT \cite{bigat}, and Social-STGCNN \cite{stgcnn} as baselines.
We reproduce these models except SoPhie and Social-BiGAT in the same environment as ours by their publicly available codes.

\begin{table*}[t]
\centering
\smallskip
\begin{tabular}{c|c|c|c|c|c|c}
\hline
\multirow{2}{*}{Model} &
\multicolumn{5}{|c|}{Dataset} & \multirow{2}{*}{Average} \\
\cline{2-6} & eth & hotel & zara1 & zara2 & univ
\\ \hline
Linear & 0.63/1.21 & 0.39/0.75 & 0.60/1.16 &    0.71/1.35 & 0.79/1.54 & 0.63/1.20 \\
CIDNN \cite{cidnn} & 1.27/1.54 & \bf{0.21/0.28} & 1.21/1.51 &   0.53/0.75 & 0.74/1.16 & 0.79/1.05 \\
SR-LSTM \cite{srLSTM} & 0.62/1.21 & 0.35/0.70 & 0.43/0.96 & 0.37/0.80 & 0.53/1.17 & 0.46/0.97 \\
SoPhie$ ^{2}$ (K=1)  & - & - & - & - & - & 0.71/1.46 \\
Social-BiGAT$ ^{2}$ (K=1) & - & - & - & - & - & 0.61/1.33 \\
\hline
Social GAN-P \cite{socialGAN} (K=20) & 0.69/1.28 & 0.48/1.02 & 0.34/0.69 & 0.31/0.65 & 0.56/1.18 & 0.48/0.96 \\
SoPhie$ ^1$ \cite{sophie} (K=20) & 0.70/1.43 & 0.76/1.67 & 0.30/0.63 & 0.38/0.78 & 0.54/1.24 & 0.54/1.15 \\
Social-BiGAT$ ^1$ \cite{bigat} (K=20) & 0.69/1.29 & 0.49/1.01 & 0.30/0.62 & 0.36/0.75 & 0.55/1.32 & 0.48/1.00 \\
STGAT \cite{stgat} (K=20) & 1.01/2.10 & 0.46/0.85 & 0.50/1.06 & 0.41/0.87 & 0.60/1.27 & 0.59/1.23 \\
Social-STGCNN \cite{stgcnn} (K=20) & 0.73/1.20 & 0.42/0.69 & {\bf 0.33/0.52} & {\bf 0.30/0.48} & 0.49/{\bf 0.91} & 0.45/{\bf 0.76}\\
\hline
BGM w/o Context Encoder (Ours, deterministic) & 0.56/1.10 & 0.30/0.58 & 0.46/0.95 & 0.34/0.71 & 0.53/1.11 & 0.44/0.89 \\
BGM w/o Social Module (Ours, deterministic) & {\bf 0.50/1.00} & 0.25/0.47 & 0.41/0.91 & 0.33/0.72 & {\bf 0.47}/1.03 & {\bf 0.39}/0.82 \\
BGM (Ours, deterministic) & 0.52/{\bf 1.00} & 0.25/0.48 & 0.43/0.93 & 0.34/0.73 & 0.48/1.03 & 0.40/0.83 \\
\hline
\end{tabular}
\caption{Quantitative results.
Results are shown as ADE/FDE in meters.
We quote results of methods with superscripts $ ^1$ from their papers, and superscripts $ ^2$ from Social-BiGAT \cite{bigat}.
}
\label{tb_ade}
\end{table*}

\subsection{Quantitative Evaluation}
\subsubsection{Compared with Other Methods}
As shown in Table.\ref{tb_ade}, SR-LSTM is a deterministic model with the best performance at present.
BGM shows more competitive performance than it:
Compared with SR-LSTM, ADE, and FDE of BGM reduce 13.0\% and 14.4\%, respectively.
Authors of \cite{bigat} provide the results of SoPhie and Social-BiGAT when generating $K=1$ trajectory.
In this case, these models can be considered as deterministic models.
For these two models, ADE of BGM has improved 43.7\% and 34.4\%, and FDE improved 43.8\% and 37.6\%, which means that BGM has a relatively stable prediction performance.

The ADE of our deterministic BGM is improved by 11.1\% compared with the best generative model Social-STGCNN that generates 20 predictions.
It should be noted that minADE of most generative models in Table.\ref{tb_ade} on UCY-zara1 and UCY-zara2 datasets have achieved a better level than deterministic models, reflecting the difference in the application purpose between these models.

Different from ADE, FDE is a significant metric that many researchers ignore.
The relative relationship between FDE and ADE reflects models' performance in predicting agents' future trends.
Visualized results (Fig.\ref{fig_vis}) shows our performance on predicting agents' future trends.
Compared with all methods listed in Table.\ref{tb_ade}, the FDE of BGM is only worse than Social-STGCNN's.
Especially, our FDE makes the best among the deterministic methods.

\subsubsection{Ablation Study}
To verify the effectiveness of each component in our model, we perform ablation experiments.
Results can be seen in the last lines in Table.\ref{tb_ade}.
\begin{itemize}
    \item Dynamic Context Encoder.
        With the context feature's help, our model could predict the agent's future trajectory, considering both his historical trajectory and the contextual information.
        Therefore, even similar trajectories will have a different prediction due to the difference in their guidance maps, reflecting the other available interaction cases between agents and the environment.
        Compared with BGM w/o Context Encoder, the ADE and FDE of BGM improve 9.0\% and 6.7\%.
    \item Social Module.
        Social interaction is always inevitable in actual sceneries.
        We use this module to make sure that the prediction meets the social rules.
        From the result of BGM w/o Social Module, its ADE and FDE have deteriorated about 2.5\% and 1.2\% compared with BGM.
        Although ADE and FDE become worse than BGM w/o Social Module, we have ensured that each prediction is sufficient to meet a safe social distance, thus avoiding possible collisions to the greatest extent and satisfying the attraction situation in the group.
        Details can be seen in Fig.\ref{fig_vis}.
\end{itemize}

\subsection{Qualitative Evaluation}
\subsubsection{Visualized Results}
\begin{figure*}[t]
\centering
\includegraphics[width=2.1\columnwidth]{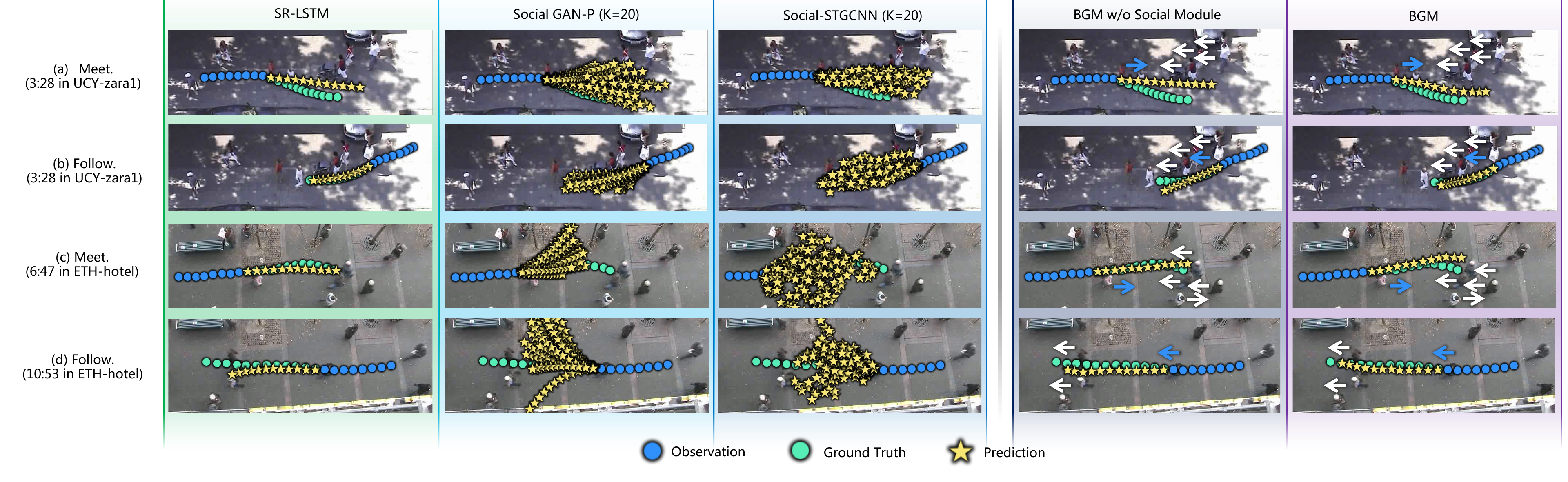}
\caption{
Visualized results of different models.
(a), (b), (c), and (d) are four different prediction conditions, which include meeting (a and c) and following (b and d) two kinds of everyday prediction situations.
Arrows indicate their walking directions.
}
\label{fig_vis}
\end{figure*}

We select several prediction scenarios to show the comparison between BGM and other existing models.
Details can be seen in Fig.\ref{fig_vis}.
We choose two kinds of ordinary prediction cases, meeting and following, to evaluate these models.
In case (a), BGM's prediction has shown a firm intention on social interaction that the target pedestrian wants to avoid others from being collided, consistent with her future behavior.
Among other methods, only SR-LSTM's prediction shows a similar trend.
Results of both Social GAN-P and Social-STGCNN seems to lack the consideration of this interaction situation.
Similar to (a), (b, c, d) show the results with specific interaction intentions.
Compared with other baselines, BGM shows a better ability to describe interactions while keeping walking styles, reflecting our strong ability to adapt to different situations.

\subsubsection{The Dynamics of Guidance Map}
\begin{figure}[t]
\centering
\includegraphics[width=0.95\columnwidth]{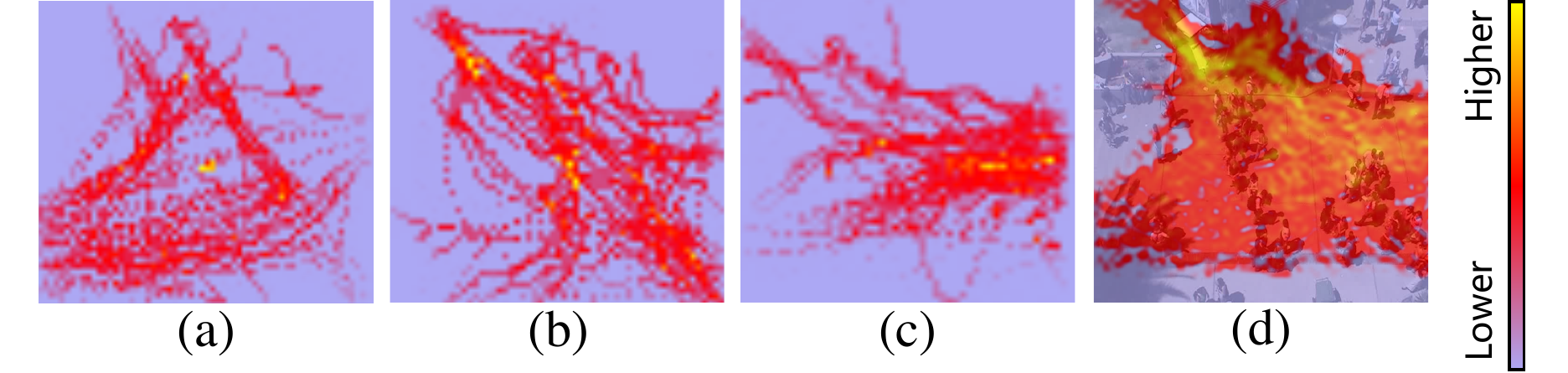}
\caption{
The dynamics of guidance map.
(a) (b) and (c) are the complete dynamic guidance maps $M(T_a)$, $M(T_b)$ and $M(T_c)$ built through three different record periods $T_a$, $T_b$ and $T_c$.
(d) shows the map generated by all available trajectories in the UCY-univ dataset.
}
\label{fig_gm}
\end{figure}

We build a guidance map for each agent in the scene to better describe both their past and future behaviors by attaching the scene context to their historical trajectories, respectively.
The guidance map's dynamic update strategy makes it available to reflect the latest trend of agents' activities in the scene.
For instance, there is a metro entrance close to a shopping mall.
Pedestrians always choose to enter the metro rather than the shopping mall in the morning when most people go to work.
However, the visual scene images or segmented images can not reflect these dynamic changes.
We visualized the complete guidance map in the UCY-univ scene in different periods in Fig.\ref{fig_gm}.
According to the update strategy shown in Algorithm.\ref{alg_timewindow}, we build three different guidance maps $M(T_a)$, $M(T_b)$ and $M(T_c)$ in three different record periods $T_a$, $T_b$ and $T_c$.
Fig.\ref{fig_gm}(d) shows the total guidance map build by all available trajectories in the whole dataset.
Fig.\ref{fig_gm}(a) (b) and (c) show different walking preferences of pedestrians in these three periods.
For example, pedestrians in $T_b$ prefer walking from right-bottom to left-top in the scene.
In contrast, most pedestrians in $T_c$ prefer to move in the horizontal direction in the scene, which shows the dynamic change of agents' flow.

We perform further comparative experiments to verify the effect of our dynamic guidance map.
As shown in Table.\ref{tb_gm}, the result shows that the BGM performs the best when the test set $X_x$ ($x\in \{a, b, c\}$) takes the local guidance map from its corresponding complete map $M(T_x)$ as the input.
(Results on the diagonal in Table.\ref{tb_gm}.)
It means that our dynamic guidance map helps the prediction, reflecting the dynamic preferences changes in the scene.

\begin{table}[t]
\centering
\smallskip
\begin{tabular}{c|c|c|c}
\hline
Test Set & $M(T_a)$ & $M(T_b)$ & $M(T_c)$ \\ \hline
$X_a$ & \bf{0.495/1.073} & 0.496/1.075 & 0.496/1.075 \\
$X_b$ & 0.437/0.937 & \bf{0.430/0.921} & 0.435/0.930 \\
$X_c$ & 0.458/0.991 & 0.463/1.004 & \bf{0.456/0.987} \\
\hline
\end{tabular}
\caption{Quantitative evaluation of the dynamics of guidance map.
We select 3 different sets of test trajectories $X_a$, $X_b$ and $X_c$ to run our prediction with their 3 corresponding complete guidance maps $M(T_a)$, $M(T_b)$ and $M(T_c)$ on UCY-univ dataset.
Results are shown in ADE/FDE.
}
\label{tb_gm}
\end{table}

\subsubsection{Social Module}
\begin{figure}[t]
\centering
\includegraphics[width=0.9\columnwidth]{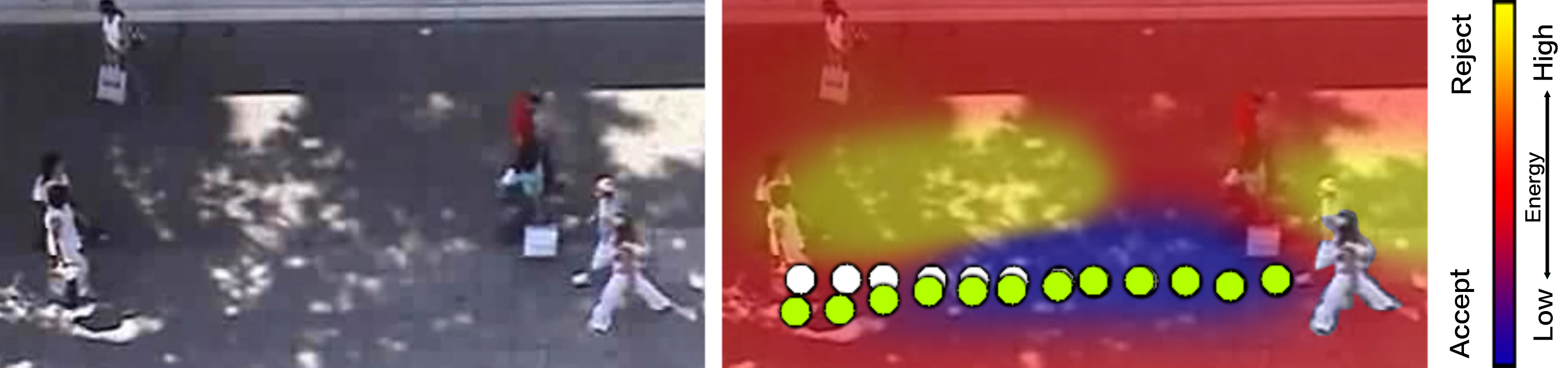}
\caption{
One social interaction case and an energy function built for the highlight pedestrian.
Little white circles represent her preliminary predictions.
Green circles show the fine-tuned prediction by the social module.
Yellow represents high energy, and blue represents low energy.
}
\label{fig_sr}
\end{figure}

The social module fine-tunes the preliminary prediction by minimizing each agent's social energy discriminant function to make their predictions socially acceptable.
This module needs not to limit the number of agents in the scene, like most current graph-based methods.
It uses a form of inferable expression, which can adapt to different situations without extra training steps.
One typical interaction case is shown in Fig.\ref{fig_sr}.
Two groups of pedestrians walking towards each other.
They will collide if they continue to walk as their historical styles.
The right figure shows the value of the pedestrian's social energy field in white clothes at each scene's position.
According to the lower energy principle, the social module suggests fine-tuning her original prediction by shifting down a little to the bottom of the scene to prevent future collisions.
The social module optimizes the preliminary predictions adaptively to a new trajectory (represented by green circles) through the lowest energy optimization, making the final prediction socially acceptable.
More interaction cases can be seen in Fig.\ref{fig_vis}.

\section{Conclusion}
This paper presents BGM for human trajectory prediction, which outperforms state-of-the-art methods on two publicly available datasets.
We construct a guidance map for each agent to explain why he behaved in the past and to guide his future movement.
It updates with the prediction period changes to adapt to the dynamic scene content.
We also propose a social module to make the predictions in line with social rules.
As shown in visualized results, BGM can give reasonable and socially acceptable predictions in complex scenarios.
Despite our contributions, there are still issues that need to be addressed.
For example, to get a better guidance map, we need to collect more available observations.
Besides, we will try considering agents' pose and intention to improve future works' prediction performance.


\bibliography{myref}
\end{document}